\documentclass[11pt]{article}
\usepackage[final]{acl}
\usepackage{booktabs}
\usepackage{siunitx}
\usepackage{makecell}
\usepackage{times}
\usepackage{latexsym}
\usepackage{todonotes}
\usepackage[T1]{fontenc}
\usepackage[utf8]{inputenc}
\usepackage{microtype}
\usepackage{inconsolata}
\usepackage[hang,flushmargin]{footmisc}
\usepackage{graphicx}
\usepackage{xcolor} 
\usepackage{tikz}
\DeclareUnicodeCharacter{03B3}{\ensuremath{\gamma}}
\newcommand{\actorcolor}{F7D0C8}
\newcommand{\actorinnercolor}{B5977A}
\newcommand{\actor}[1]{%
    \setlength{\fboxsep}{1pt}%
    \colorbox[HTML]{\actorcolor}{#1}}
\newcommand{\actorinner}[1]{%
    \setlength{\fboxsep}{1pt}%
    \colorbox[HTML]{\actorinnercolor}{#1}}

\newcounter{myex}
\makeatletter
\newcommand{\myex}[1]{%
    \refstepcounter{myex}%
    \par\vspace{2pt}%
    {\linespread{1.0}\selectfont%
    \leftskip=0em\rightskip=1em%
    \hangindent=1.8em\hangafter=1%
    \small\noindent(\arabic{myex})\hspace{0.5em}#1\par}%
    \vspace{2pt}%
    \@afterindentfalse\@afterheading%
}
\makeatother
\newcommand{\myref}[1]{{\small\ref{#1}}}

\usepackage{needspace}

\makeatletter
\renewcommand{\paragraph}{%
    \@startsection{paragraph}{4}{\z@}%
    {1.5ex \@plus 0.5ex \@minus .2ex}%
    {-0.7em}%
    {\normalfont\normalsize\bfseries}%
}
\makeatother

\title{Who's Behind It? Annotating and Extracting Conspiratorial Actors from German Telegram Posts}

\author{Helena Mihaljevi\'{c} \\\and {\bf Jolanda Beer} \\\and {\bf Mareike Lisker} \\
         HTW Berlin, Germany\\
                 \texttt{mihalje@htw-berlin.de}
         \And Katharina Soemer \\ Goethe University Frankfurt, Germany\\
        }

\begin{document}
\maketitle
\begin{abstract}

Conspiracy theories commonly attribute important events to the actions of powerful and secretive actors. While computational research has largely focused on document-level analyses of conspiracy theories, less attention has been paid to identifying the actors portrayed as responsible in such narratives. We develop detailed guidelines for the span-level annotation of conspiratorial actors and use them to create a corpus of German Telegram posts, before investigating automatic actor extraction using transformer-based models. We further apply the resulting model to the \textit{Schwurbelarchiv}, a large-scale archive of German conspiracy-related Telegram channels. Our results demonstrate that conspiratorial actors can be annotated consistently and extracted with reasonable accuracy despite the linguistic complexity of conspiracy discourse, enabling large-scale analyses of actor representations in conspiracy narratives.

\end{abstract}

\section{Introduction}

Conspiracy theories (CTs) are commonly understood as narratives that explain important events through the intentional actions of powerful and secretive actors pursuing allegedly malicious goals \cite{butterConspiracyTheoriesConspiracy2021,douglasWhatAreConspiracy2023}. They are a recurring form of sense-making in public discourse, typically gaining prominence during periods of uncertainty and crisis, and affecting trust in democratic institutions, public health behavior, and the interpretation of societal events \cite{samoryConspiraciesOnlineUser2018,vanprooijenConspiracyTheoriesPart2017,douglasUnderstandingConspiracyTheories2019}.  
Moreover, CTs frequently promote dehumanizing, antisemitic, racist or otherwise hateful worldviews and have been associated with hate speech and far-right radicalization \cite{vergani2022,schulze_far_right_2022}.

Computational research on conspiracy theories has grown substantially in recent years. Numerous datasets and computational approaches have been proposed for CT detection, stance classification, topic analysis, or the study of information diffusion \cite{moffittHuntingConspiracyTheories2021,steffen_etal_icwsm_2023,langguthCOCOAnnotatedTwitter2023,ACTI_EVALITA_2023,corso_ct_tiktok_2025,pustet_detection_2024,liuConspEmoLLMConspiracyTheory2024,steffen_more_2025}. While these approaches have improved the identification of conspiracy-related content, they typically represent CTs as document-level labels. 
As a result, relatively little is known about how conspiracy narratives themselves are structured and communicated at scale.

Recent work has begun to model recurring narrative components  \cite{batzdorfer_conspiracy_2024,piva_conspiracy_2026} or cognitive traits \cite{bates_conspired_2025} within conspiratorial content.
Among these components, actors occupy a central role. By identifying allegedly hidden actors behind seemingly unrelated events, conspiracy theories transform coincidence into intention and construct coherent explanatory narratives \cite{barkun_culture_2003}.  
The PsyCoMark shared task introduced the automatic span-level extraction of five psycholinguistic markers of conspiracy narratives \cite{samory_semeval_2026}. Although actors were among the more reliably identified markers, both annotation agreement and system performance indicate substantial room for improvement.

We argue that reliable conspiratorial actor extraction requires addressing several challenges: actor mentions can be collective, implicit, highly descriptive, or identifiable as actors only from context, making their identification fundamentally different from conventional named entity recognition. This calls for a careful operationalization of conspiratorial actors and detailed annotation instructions that account for the semantic and syntactic ambiguities involved in their identification and delimitation.
Reliable extraction, in turn, is a prerequisite for systematically studying conspiratorial actors at scale. Accordingly, little is known about which types of actors are constructed in conspiracy discourse and how their representation changes over time.

We address these gaps by developing detailed annotation guidelines for conspiratorial actors and applying them to create a span-annotated corpus of German Telegram posts. Building on this corpus, we train and evaluate transformer-based actor extraction models. Finally, we demonstrate their utility for large-scale analysis by examining the representation and temporal evolution of conspiratorial actors in the \textit{Schwurbelarchiv} \cite{angermaier_schwurbelarchiv_2026}, a German Telegram archive associated with conspiracy-theory communication during the COVID-19 pandemic.

\section{Related Work}

Viewing conspiracy theories as  sense-making narratives \cite{fenster_conspiracy_2008}
has inspired computational approaches that model conspiracy narratives through entity-relation networks or automatically induced narrative graphs \cite{shahsavariConspiracyTimeCorona2020,tangherlini_automated_2020}. 
Several studies have proposed structured representations of conspiracy narratives. \citet{tangherlini_automated_2020} reconstruct corpus-level narrative graphs from automatically extracted actants and their relations, while \citet{batzdorfer_conspiracy_2024} operationalize conspiracy discourse through features such as actors, actions, threats, secrecy, and patterns. \citet{piva_conspiracy_2026} introduce a span-level frame representation comprising plans, secrets, in-groups, out-groups, and calls to action. In a related line of work, \citet{bates_conspired_2025} annotate CT texts for cognitive traits such as overriding suspicion and malicious intent.

In parallel to our work, the SemEval-2026 PsyCoMark shared task introduced the span-level annotation and extraction of five psycholinguistic markers of conspiracy narratives, including actors \cite{samory_semeval_2026}. The dataset was annotated by crowdworkers using a compact codebook that defines actors through their alleged responsibility for a malicious action or agenda.\footnote{PsyCoMark annotates markers independently of document-level
conspiracy classification, including in texts classified as non-conspiratorial.}
The authors report limited agreement for marker spans overall (unitizing Krippendorff's $\alpha=0.15$; macro-F1 $=0.19$ at IoU $\geq 0.5$). Actor was among the best-performing markers in the subsequent extraction task, with an overlap-based F1 of 0.28 for the baseline model and approximately 0.40 for the best-performing systems.

Taken together, existing work demonstrates the value of representing conspiracy narratives through their constituent components, but the detailed operationalization of conspiratorial actors for consistent span-level annotation and automatic extraction remains underexplored. Moreover, while actors and their relations have been featured in previous studies of conspiracy narratives \cite{tangherlini_automated_2020}, their systematic analysis across large-scale conspiracy discourse has received limited attention. Our work addresses these gaps by developing and evaluating a detailed framework for conspiratorial actor annotation and extraction and using it to systematically analyze actor representations at scale.

\section{Actor Annotation}

\subsection{Conspiratorial Actors}

We define \textit{conspiratorial actors}, or \textit{actors} for short, as individuals, groups, organizations, or institutions---real or imagined---that conspiracy narratives portray as intentionally responsible for societal events through the pursuit of allegedly malicious goals. Their linguistic realization ranges from specific names, e.g., of individuals or organizations, to more or less abstract collectives, pronouns, or implicit references, making actor identification substantially more challenging than conventional named-entity recognition (NER). We include all entities portrayed as intentionally advancing the alleged conspiracy, including both principal actors and supporting actors.

\subsection{Corpus}

We extend the \textit{TelCovACT} dataset \cite{steffen_etal_icwsm_2023}, a collection of German-language Telegram posts from conspiracy-oriented channels collected between 03/2020 and 12/2021. The original dataset was annotated at the document level for the presence of a CT and  the narrative components \textit{actor}, \textit{strategy} and  \textit{goal}. For the present study, we extend the subset of 1,251 posts labeled as propagating a CT (out of 3,663 posts in the full dataset) with span-level annotations of \textit{actor} mentions, the most frequent narrative component in the dataset.

In the final corpus\footnote{The annotated dataset is available upon request via Zenodo: \url{https://doi.org/10.5281/zenodo.22644678}.}, 922 posts (73.7\%)\footnote{The original \textit{TelCovACT} document-level annotations flag the presence of actors in 801 of the 1,251 CT-positive posts. For the present study, all 1,251 posts were independently inspected for actor spans, irrespective of their original actor label. Disagreements in the span-level annotations were discussed among the annotators, while the original document-level labels were left unchanged. Consequently, the two annotation layers do not necessarily coincide.} 
contain at least one actor span. The distribution of annotated spans per post is strongly right-skewed, with a mean of 3.45 spans but 338 posts (36.7\%) containing a single span, 215 posts (23.3\%) containing two, and 113 (12.3\%) containing three. The number of spans is only moderately correlated with text length (Spearman's $\rho = 0.36$). Actor spans are generally short, with  three characters being the most frequent span length, largely reflecting pronominal references such as `sie' (`they', `she') and abbreviations such as `WHO' or `USA'.

\subsection{Annotation Guidelines and Process} 
Conspiratorial actors constitute a challenging target for span annotation. 
Unlike named entities, they are not defined by entity type or unique real-world reference, but by the role they assume within a conspiracy narrative. Moreover, actor mentions may be  abstract or concrete, explicit or implicit, and real or fictional. Unlike semantic roles in event argument extraction (EAE), conspiratorial actors are not tied to a specific lexical trigger but assume their narrative role relative to the conspiratorial meaning of the text as a whole. 
In EAE, multi-word arguments can be annotated through their syntactic head words and subsequently expanded to complete spans using dependency parsing \cite{zhang_transfer_2022}. We instead annotate the \textit{smallest self-contained phrase identifying the conspiratorial actor}, as it facilitates later analyses of actor representations while remaining largely independent of syntactic structure.

Our guidelines thus build on established span annotation concepts from NER and EAE while tailoring them to the specific properties of conspiratorial actors. Since determining exact span boundaries is inherently ambiguous \cite{feng_probing_2020}, with annotators often agreeing on the relevant semantic region but not on its exact extent \cite{lee_study_2019}, our guidelines explicitly address syntactic decisions involved in actor span delimitation.

The guidelines were developed  iteratively through repeated annotation rounds, jointly discussing  disagreements and refining the rules throughout the process. The complete annotation guidelines are provided in Appendix~\ref{sec:app_annotation_guidelines}.

The annotation was conducted by two authors of this paper: an undergraduate student who was compensated for the annotation work and a senior researcher in data science and NLP with prior experience in annotating CT data. The student annotator received training that included literature on conspiracy theories, existing guidelines for document-level CT annotation, and several rounds of test annotation followed by discussion.
Each post was initially annotated by one annotator. Ambiguous cases were flagged during annotation and jointly reviewed, with final annotations determined through discussion.

\subsection{Inter-Annotator Agreement}
\label{sec:interannotator_agreement}
To assess annotation reliability, both annotators independently annotated an additional sample of 100 posts from a separate collection of conspiracy-theory Telegram channels from a similar time period.
Agreement was measured using Gamma \cite{mathet_unified_2015}, a span-aware agreement measure for unitizing tasks, together with strict span-level F1 (SeqEval), relaxed F1 counting overlapping spans as matches, and character-overlap F1 measuring agreement over the character positions covered by actor spans.
Agreement scores between 0.73 and 0.79 indicate that conspiratorial actors can be annotated consistently despite the inherent ambiguity of the task. 
Detailed results are presented in Table~\ref{tab:appendix_iaa} in Appendix \ref{sec:appendix_iaa}.

Only 17\% of disagreements involved span boundaries, whereas 83\% concerned whether an expression was considered an actor at all.
To better understand these disagreements, we manually inspected all inconsistent annotations. 
Most presence/absence disagreements were straightforward annotation omissions, where the annotators agreed upon review that an actor mention had been overlooked. These omissions involved especially personal and impersonal pronouns but also explicit actor mentions, and occurred both for first mentions and for repeated references to actors already annotated earlier in the post. Other presence/absence disagreements reflected semantic differences in interpreting the role of an expression within the conspiracy narrative. These mainly concerned nested actors, helper versus affected groups, country names functioning as actor specifications or geographical references, and descriptive or metaphorical actor expressions.
Span boundary disagreements resulted from differences in the inclusion of modifiers, coordinated noun phrases, appositions, or other syntactic constituents. 

\section{Actor Extraction}
\label{sec:actor_extraction}

Actor extraction was formulated as a BIO-based token-classification task \cite{ramshaw_text_1995}.
We fine-tuned German transformer models using the HuggingFace Transformers framework and evaluated multiple hyperparameter configurations varying learning rate, batch size, and weight decay. Hyperparameters were selected using stratified 5-fold cross-validation. The selected configuration was then retrained on the combined training and validation data using five random seeds and evaluated on the held-out test set. Model performance was assessed using strict, relaxed and overlap-based F1, as introduced in Section \ref{sec:interannotator_agreement}. In addition to the token-level evaluation, we report corresponding word-level metrics obtained by projecting token predictions back to complete words.

Across five random seeds, the selected model achieved a mean strict span-level F1 score of 0.52 on the held-out test set. Relaxed and overlap-based metrics reached around 0.65, indicating that span boundary errors account for a substantial part of the  extraction errors. Word-level evaluation produced comparable results. 

Our results indicate that automatic actor extraction is sufficiently reliable to support large-scale analyses of conspiracy narratives. Detailed training settings and evaluation results are provided in Appendix \ref{sec:appendix_actor_extraction}.

\section{Narrative Actors in \textit{Schwurbelarchiv}}

We demonstrate the usefulness of automatic actor extraction by applying our model to the  \textit{Schwurbelarchiv}, a large-scale archive of German-language Telegram channels associated with conspiracy-theory communication, kindly provided to us by the authors of \cite{angermaier_schwurbelarchiv_2026}. We restrict the analysis to unique, non-transcribed conspiracy-theory posts of at most 2,000 characters, consistent with both the preceding CT detection in the \textit{Schwurbelarchiv} and our transformer-based extraction setting, yielding $\sim\!\!1.35$ million posts. 

Actor extraction identified 1,878,993 actor spans, with 918,353 posts (67.8\%) containing at least one actor (1.39 spans per CT post on average). Actor spans were normalized by lowercasing and removing whitespace and punctuation, yielding 268,897 distinct actors. The large number of unique actors reflects the long-tailed distribution of conspiratorial actors.
Further corpus details are provided in Appendix~\ref{sec:appendix_schwurbelarchiv}.

In addition to the observed large variety of normalized actors, a manual inspection of the extracted actors provides evidence of  generalization beyond actors seen in training data, including entities associated with unseen events and geopolitical contexts such as the Russian invasion of Ukraine. Moreover, an analysis of the 20 most frequent extracted actors shows that the model does not simply label typical lexical expressions as actors whenever they occur; rather, only a proportion of their occurrences are identified as actors, and this proportion varies substantially across expressions. For example, `those' (German: `die') is identified as an actor in only a small proportion of its occurrences, whereas `freemasons' (German: `Freimaurer') is almost always identified as an actor.

\subsection{Referential Specificity of Actors}

To characterize how actors are referenced, we classify extracted spans according to their referential specificity (Table~\ref{tab:actors_schwurbelarchiv}). The taxonomy distinguishes four levels ranging from implicit references to uniquely identifiable entities.
Two authors jointly assigned the most frequent extracted actors to one of the four abstraction levels and labeled whether those extracted actors reference to  antisemitic, racist, or sexist  narratives. Moreover, typical naming variants (e.g., `mainstream media', `mass media'  and `MSM') were identified, resulting in 1,075 manually categorized spans. Subsequent normalization of frequent spelling variants, inflectional forms, and highly similar surface forms extended the categorization to ~63,3\% of all extracted spans, which we refer to  as \textit{annotated actors}. 

\begin{table}[h]
\centering
\small
\caption{Actor taxonomy based on  referential specificity and prevalence in the \textit{Schwurbelarchiv}.}
\label{tab:actors_schwurbelarchiv}
\begin{tabular}{cllr}
\toprule
\textbf{L} & \textbf{Category} & \textbf{Examples} & \textbf{Share} \\
\midrule
1 & Implicit  & they, he, those & 19.1\%\\
2 & Abstract  & elites, satanists, globalists &22.6\% \\
3 & Concrete  & media, doctors, military & 29.4\%\\
4 & Specific  & WHO, George Soros, Pfizer  & 28.9\%\\
\bottomrule
\end{tabular}
\end{table}

Among the annotated actors, concrete collectives form the largest category, closely followed by specific entities (see Table~\ref{tab:actors_schwurbelarchiv}). However, more than 40\% of actor spans are either abstract or completely implicit. The overall proportion of implicit actors is likely overestimated, as we expect the annotation of implicit references to be close to exhaustive. Even under the conservative assumption that none of the non-annotated actor mentions are implicit, 8.5\% of all actor-containing posts refer exclusively to implicit actors. This suggests that a substantial portion of conspiracy narratives rely on shared background knowledge without explicitly naming the alleged perpetrators. 

Among the annotated ideological tropes, actors referencing antisemitic conspiracy narratives are by far the most prevalent. They account for 15.7\% of all spans and occur in 15.3\% of posts containing at least one actor. In contrast, misogynistic references are considerably less frequent (4.8\% of spans; 5.0\% of posts), while racialized actors play only a marginal role (0.5\% of spans and posts).

The prevalence of actor-containing posts largely follows the overall temporal distribution of conspiracy-theory posts in the corpus (Figure~\ref{fig:actor_distribution} in Appendix~\ref{sec:appendix_schwurbelarchiv}). Across the observation period, most actor-containing posts include at least one non-implicit actor, while posts referring exclusively to implicit actors are considerably less frequent.

\begin{figure}[h]
    \centering
    \includegraphics[width=0.999\linewidth]{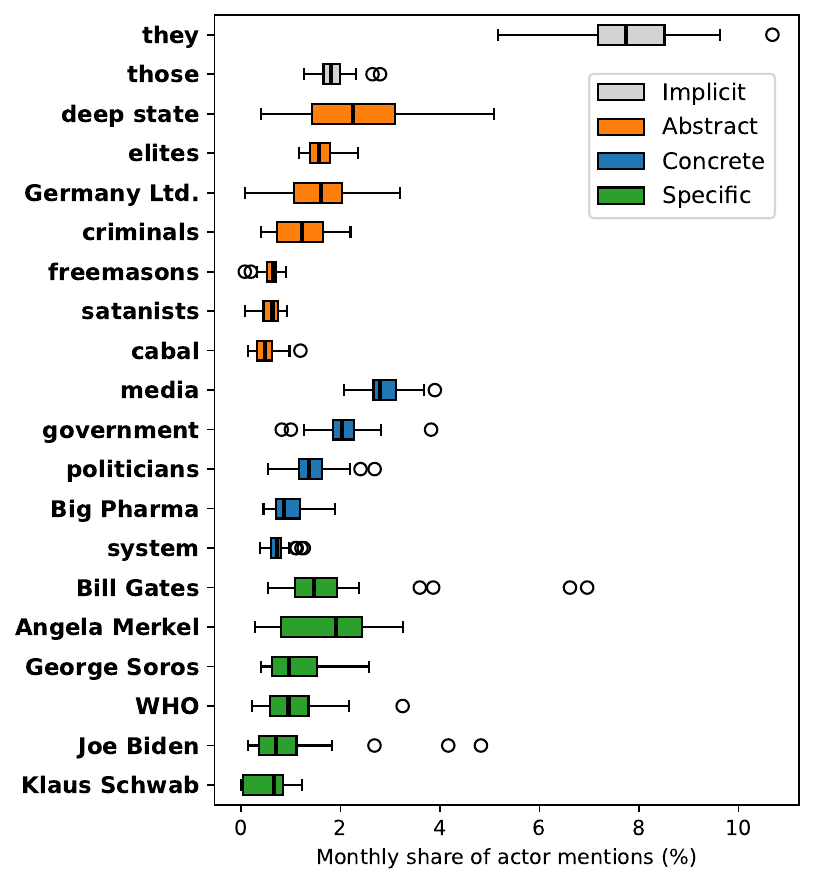}
\caption{Monthly distribution of the 20 most frequent normalized conspiratorial actors.}   
\label{fig:top_actors}
\end{figure}

Among concrete actors, `media', `government', and `politicians' remain comparatively stable over time, as indicated by the small boxes in Figure~\ref{fig:top_actors}. Specific actors show higher variance, suggesting event-related prominence. Abstract actors show mixed behaviour: while `freemasons' and `cabal' remain consistently present, pointing to a stable antisemitic core in the \textit{Schwurbelarchiv}, `deep state' and `Germany Ltd.'\footnote{Refers to a CT claiming that Germany is not a sovereign state but a corporation established after WW II.} vary considerably. The most frequent actor overall, the implicit reference `they'\footnote{German \textit{sie}/\textit{Sie} can refer to `they'/`them', `she'/`her', or formal `you'. We collapse these forms and render them as `they' in Figure \ref{fig:top_actors}, since manual inspection of a sample suggests that the plural meaning is the most frequent in our data.}, ranges from roughly 6\% to over 10\% of monthly actor mentions.

\section{Discussion and Conclusion}

Conspiratorial actors differ fundamentally from conventional named entities because they are frequently abstract, implicit, or only partially specified. Nevertheless, our annotation achieves substantial inter-annotator agreement, and transformer-based models extract such actors with reasonable accuracy. This demonstrates that actor extraction is a feasible building block for computational analyses of conspiracy narratives.

The proposed taxonomy provides an analytical lens for studying how conspiracy narratives refer to alleged perpetrators across different levels of specificity. 
Applying the model to more than one million conspiracy theory posts from the Telegram-derived \textit{Schwurbelarchiv} reveals a highly diverse, long-tailed actor distribution together with a persistent core of actors that remains prominent across different societal crises, while event-specific actors seem to emerge and disappear over time. Moreover, antisemitic actors constitute a central and persistent component in German-language conspiracy communities represented in the corpus.
Their high prevalence relative to actors associated with misogynistic or racist narratives may partly reflect the corpus's temporal focus on the COVID-19 pandemic, as other political contexts, such as elections, can foreground different forms of prejudice \cite{simmons_gendered_disinfo_2024}. At the same time, the projective character of antisemitism \cite{horkheimer_adorno_2002} may help explain the prominence of antisemitic actors in narratives surrounding otherwise unrelated societal crises and the adaptability of antisemitic stereotypes to changing contexts.

Overall, our results demonstrate that actor extraction enables analyses that go beyond document-level conspiracy detection and provide insights into the structure and evolution of conspiracy narratives at scale. Future work will extend to additional narrative components and investigate their relations, enabling a more comprehensive computational reconstruction of conspiracy narratives.

\section{Limitations}

Our annotation guidelines and corpus are currently limited to German-language Telegram posts. While the concept of conspiratorial actors is likely to generalize across languages and platforms, their linguistic realization differs across languages. Compared with English, which typically favors more verbal and syntactically unpacked constructions, German tends to express concepts through dense nominal phrases, compound nouns, nominalizations, and richer inflectional morphology. These differences may affect both span boundaries and annotation decisions, motivating future validation and adaptation of the guidelines for other languages.

Our annotation task is performed on individual posts. Some actor mentions labeled as implicit may therefore rely on contextual information from preceding posts or conversations. Incorporating conversational context should thus be addressed in  future work.

Finally, our large-scale analysis is conditioned on the composition of the \textit{Schwurbelarchiv} and may also inherit biases and errors from the preceding automatic CT detection. The observed actor distributions should therefore not be assumed to generalize to German-language conspiracy discourse more broadly. Moreover, the analysis relies on encoder-based transformer models and short textual posts. In preliminary experiments, LLMs frequently inferred plausible conspiratorial actors that were not explicitly mentioned in the text, despite explicit prompting instructions to extract only textual spans. Future work should investigate how this tendency can be mitigated, as it would enable reliable span extraction from substantially longer texts, including automatically transcribed audio and video content, which remains largely unexplored in conspiracy-theory research.

\bibliography{references}

\newpage
\appendix

\section{Annotation Guidelines}
\label{sec:app_annotation_guidelines}
Below we present the annotation guidelines. Although the texts to be annotated are in German and several rules are specific to properties of the German language, the annotation rules and examples are provided in English to ensure broader accessibility and to facilitate reuse in future annotation efforts. Where necessary to illustrate the specificity of a German construction, examples are provided in German with an English translation. Our examples contain either short phrases or entire sentences to illustrate the rules.

\setlength{\parskip}{1pt}
\needspace{3\baselineskip}
\subsection{Background}
The central goal of this document is to facilitate the annotation of the \textbf{actor} narrative component in German-language online and social media texts containing conspiratorial content.
We follow \citet{butter_nichts_2018} in defining conspiracy theories as narratives premised on the strong belief that a secret group of people intentionally causes complex, and in most cases unsolved, events in pursuit of an evil goal, such as taking over an institution, a country, or even the entire world. Such narratives are commonly characterized by four recurring narrative components \cite{steffen_etal_icwsm_2023,batzdorfer_conspiracy_2024}: a malicious \textit{actor} (e.g. corrupt elites), who enacts a \textit{strategy} (e.g. inserting a microchip via vaccinations) in pursuit of a \textit{goal} (e.g. controlling the population), all while maintaining \textit{secrecy}.

The annotation rules operate on three levels. At the \textbf{semantic level}, rules determine which entities qualify as a malicious actor within a conspiratorial narrative. 
Where ambiguities arise and boundaries are fuzzy, the \textbf{syntactic level} provides additional resolution. Syntax-level rules are broadly informed by X-bar theory \citep{chomsky_remarks_1970, jackendoff_x_1977}, which models phrases as hierarchical structures built around a lexical head. Concretely, we include words and phrases that modify the head noun in a nominal phrase, such as adjectives and prepositional phrases, while excluding elements that are structurally more peripheral, such as relative clauses, as well as elements outside the nominal phrase entirely, such as verbs and determiners. How these principles apply in practice will become apparent throughout the annotation rules below. 
At the \textbf{character level}, rules govern the treatment of surface forms such as punctuation, quotation marks, and emojis.

Note that while some rules provide explicit, determinate criteria, others are better understood as heuristics reflecting recurring patterns in the data. If an annotator is unsure about a labeling decision, they should tick the ``to discuss'' box for later joint assessment.

\needspace{3\baselineskip}
\subsection{Rules for Annotating the Actor}

\needspace{3\baselineskip}
\subsubsection{Semantic Level}

\needspace{3\baselineskip}
\paragraph{Definition.} We label an entity as actor if they are mentioned or implied as supposedly and maliciously involved in a conspiracy (\myref{ex:pharmacies_actor}).

\myex{\label{ex:pharmacies_actor} \actor{Pharmacies} push experimental vaccines on us!}

Entities can be existing or non-existing individuals (\myref{ex:gates}, \myref{ex:satan}), or more or less concrete groups of people, organizations or institutions (\myref{ex:who}, \myref{ex:government}, \myref{ex:elites}, \myref{ex:satanists}).

\myex{\label{ex:gates} \actor{Bill Gates} is funding the global vaccination agenda to control us!}
\myex{\label{ex:satan} \actor{Satan} himself is pulling the strings behind the NWO.}
\myex{\label{ex:who} The \actor{WHO} is withholding the truth about the dangers of vaccines.}
\myex{\label{ex:government} We have been lied to about the true purpose of the lockdowns by the \actor{government}.}
\myex{\label{ex:elites} This crisis was orchestrated from the very beginning. Who else but the \actor{elites}?}
\myex{\label{ex:satanists} Our children are being corrupted by \actor{satanists} in Hollywood!}

Personal pronouns in the nominative case referring to an actor are annotated as well (\myref{ex:explicit}).

\myex{\label{ex:explicit} \actor{They} are watching us.}

Note that the guidelines assume that the annotated post propagates a conspiracy narrative. Posts that have been mistakenly labeled as conspiracy-related should not have their entities annotated as actors, even if those entities frequently appear in conspiratorial contexts (see negative examples at the end of the guidelines).

\needspace{3\baselineskip}
\paragraph{Basis for annotation.} Annotation decisions are based solely on the content of the post, which thus should be read entirely first of all. Annotators should not guess or infer beyond what is explicitly stated, thus implicit references in passive constructions are not sufficient for annotation (\myref{ex:implicit}).

\myex{\label{ex:implicit} We are being watched.}

However, drawing on prior knowledge of well-known conspiracy theories and figures is legitimate and encouraged where relevant (\myref{ex:q}).

\myex{\label{ex:q} As Q said, \actor{they} have started.}

As the content of the entire post forms the basis for annotation decisions, even if the role of an alleged actor becomes clear only later in the post, they are marked as an actor in every instance in which they are mentioned (\myref{ex:neutral-first}).

\myex{\label{ex:neutral-first} The \actor{WHO} published a new report on vaccine safety. [...] Recent investigations revealed that the \actor{WHO} has been hiding data on severe side effects for years!}

The only exception applies when an actor is first mentioned in a clearly positive light and only later revealed to be malicious. In this case, the initial positive mention is not annotated (\myref{ex:friend-canderian}).

\myex{\label{ex:friend-canderian} I forwarded this question to a friend of mine, Dr. Mylo Canderian. [...] To be  transparent, \actor{Dr. Canderian} is involved in the vaccination genocide.}

\needspace{3\baselineskip}
\paragraph{Scope.} The actor label also covers secondary enablers acting in service of the primary alleged orchestrators (\myref{oelrich0}, \myref{ex:doctor}).

\myex{\label{oelrich0} \actor{Stefan Oelrich} works for the \actor{Bayer AG} to orchestrate the rollout of the mRNA vaccine to fulfil the globalist depopulation agenda.}
\myex{\label{ex:doctor} \actor{Doctors} across the country are knowingly administering the poisonous vaccine on behalf of \actor{Big Pharma}.}

However, not every grammatical subject that performs an action should be labeled as actor, even if it appears in a conspiracy-related context (\myref{ex:pharmacies_noactor}, \myref{ex:polizisten}).

\myex{\label{ex:pharmacies_noactor} Pharmacies are to play a key role in the vaccination effort.}
\myex{\label{ex:polizisten} What else has to happen before even the last \actor{corrupt police officer} realizes that he is a \actor{puppet of the \actorinner{regime}}? How do police officers behave when the people rise up and have had enough of the violence coming from the government and enforced by the police?\\\textit{Note that the second usage of ``police officer(s)'' is not annotated.}\\\textit{For the nested annotation of ``puppet of the regime'', see the paragraph on Nested actors.}}

\needspace{3\baselineskip}
\paragraph{Comparisons.} Entities used purely as a point of comparison to characterize an actor's behavior are not themselves annotated as actor, even if the compared entity is itself malicious (\myref{ex:nazis}).

\myex{\label{ex:nazis} \actor{They} are like the National Socialists of back then.}

\needspace{3\baselineskip}
\paragraph{Geographic references.} Country names are only annotated as actor when they imply agency or culpability rather than serving as a mere geographic reference (\myref{ex:deutschland}, \myref{ex:israel}).

\myex{\label{ex:deutschland} \actor{Germany} has exported the \actor{Antifa} to the US.}
\myex{\label{ex:israel} Number of Deaths from Vaccine-Related AIDS in Israel Overwhelms Funeral Directors [...] \actor{they} achieve exactly what \actor{they} want to achieve.}

\paragraph{Predicate nominatives.} Predicate nominatives, i.e. noun phrases that are linked to a subject via a copular verb such as \textit{to be} (``sein'') or \textit{to become} (``werden''), are annotated if they denote a malicious role (\myref{ex:traitors}, \myref{ex:mercenaries}).

\myex{\label{ex:traitors} \actor{They} are \actor{traitors}.}
\myex{\label{ex:mercenaries} We no longer have a police force, \actor{they} are \actor{paid mercenaries} now.}

\subsubsection{Syntactic Level}

\needspace{3\baselineskip}
\paragraph{Span boundaries.} We include only elements directly relevant to the actor's role within the conspiracy narrative. Thus, in line with X-bar theory and to avoid including irrelevant or complex adjectival phrases, determiners such as definite or indefinite articles are not included in the span (\myref{ex:who}, \myref{ex:government}).

Adjectival phrases and other modifiers are included only if conspiracy-relevant (\myref{ex:corrupt-chancellor}, \myref{ex:covid-restrictions}), and excluded if purely descriptive (\myref{ex:old}).

\myex{\label{ex:corrupt-chancellor} The \actor{corrupt chancellor}, who announces new measures, is plotting against the people.}
\myex{\label{ex:covid-restrictions} The \actor{proponents of the dangerous COVID-19 restrictions}\\\textit{Note that the adjunct is a prepositional phrase in English, whereas in German it is a nominal phrase: ``Die \actor{Befürworter \underline{der gefährlichen}} \actor{\underline{Covid-Verschärfung}}''.}}
\myex{\label{ex:old} the old \actor{chancellor Helmut Schmidt}}

This applies although the words in the span would be ungrammatical in isolation, as is often the case in German where nouns are declined according to grammatical case (\myref{ex:illuminaten_article}).

\myex{\label{ex:illuminaten_article} Die \actor{mächtigen Illuminati} kontrollieren die Weltpolitik.\\\textit{The \actor{mighty Illuminati} control world politics.}}

Further, verbs are also not included in the span, even when they provide additional information about the actor, since noun phrases are embedded within the verb phrase rather than the other way around (\myref{ex:explicit}, \myref{oelrich0}). Relative clauses are likewise excluded from the span, although in X-bar theory they would be embedded within the nominal phrase.

In genitive constructions, the opening article is omitted from the span (\myref{ex:genitiv_tief}), while the genitive suffix is annotated (\myref{ex:genitiv_merkel}), which also follows from the sub-word annotation rule below.

\myex{\label{ex:genitiv_tief} Rückgrat des \actor{tiefen Staats}\\\textit{The backbone of the \actor{deep state}}}
\myex{\label{ex:genitiv_merkel} \actor{Merkels} plan}

Regarding nominalized adjectives, where the article is obligatory as a nominalization marker, it is included in the span (\myref{ex:böse}).

\myex{\label{ex:böse} Volksidealismus ist \actor{dem Bösen} Hindernis.\\\textit{Popular idealism is an obstacle to \actor{evil}.}}

While possessive pronouns as determiners are generally not included in the span, an exception applies when the possessive pronoun is integral to identifying the group and omitting it would render the span uninterpretable (\myref{ex:unsere_rasse}).

\myex{\label{ex:unsere_rasse} ``\actor{Our race} is the master race [...] \actor{We Jews} are the destroyer [...]''}

\needspace{3\baselineskip}
\paragraph{Name and role.} When a proper name appears together with a job title or role description in one phrase, i.e. because one nominal phrase is adjunct to the other, both are annotated as a single span, regardless of their order in the text (\myref{ex:fake-president}, \myref{ex:dr-canderian}). Honorifics (e.g. \textit{Mr.}, \textit{Mrs.}) are likewise included in the span (\myref{ex:mrs-clinton}). Note that there may be nested annotations in consideration of the following Nested actors rule.

\myex{\label{ex:fake-president} \actor{fake president Biden}}
\myex{\label{ex:dr-canderian} \actor{Dr. Canderian, a medical staff member} \actor{at the \actorinner{WHO}}}
\myex{\label{ex:mrs-clinton} \actor{Mrs. Clinton}}

We have two separate annotations if the proper name and the role description appear as two objects or separate noun phrases in a sentence (\myref{ex:barroso}).

\myex{\label{ex:barroso} \actor{José-Manuel Barroso} has now been appointed board member of the \actor{vaccine organisation \actorinner{GAVI}}.}

Note that in German, constructions such as in (\myref{ex:chef-reardon}) are possible, where it may appear as though \textit{Chef Tim Reardon} should be annotated as a single span. However, within the subordinate clause, \textit{Chef} and \textit{Tim Reardon} constitute two independent nominal phrases and are therefore annotated independently.

\myex{\label{ex:chef-reardon} [...], deren \actor{Chef} \actor{Tim Reardon} ist.\\\textit{[...], whose \actor{boss} is \actor{Tim} \actor{Reardon}.}}

\needspace{3\baselineskip}
\paragraph{Nested actors.}\label{rule:nested} Actors can be nested, i.e., one actor can be embedded within another, reflecting the hierarchical structure of conspiracy narratives in which secondary enablers act on behalf of a larger conspiratorial entity (\myref{ex:putin_puppets}, \myref{ex:greta}).

\myex{\label{ex:putin_puppets} \actor{\actorinner{Putins} puppets}}
\myex{\label{ex:greta} \actor{\actorinner{Greta} follower}}

Syntactically, this embedding can take different forms: the inner actor may appear as a modifier within a compound (\myref{ex:board_member}), as a separate nominal phrase adjunct (\myref{ex:oelrich_adjunct}), or as an appositive construction (\myref{ex:oelrich_apposition1}, \myref{ex:oelrich_apposition2}).

\myex{\label{ex:board_member} \actor{\actorinner{Bayer} board member}}
\myex{\label{ex:oelrich_adjunct} \actor{board member of \actorinner{Bayer AG}}}
\myex{\label{ex:oelrich_apposition1} \actor{board member of the} \actor{\actorinner{Bayer AG} Stefan Oelrich}}
\myex{\label{ex:oelrich_apposition2} \actor{Stefan Oelrich, board member} \actor{of the \actorinner{Bayer AG}}}

\paragraph{Enumerations.} When enumerated, we label each actor individually (\myref{ex:enumeration}, \myref{ex:illuminaten}).

\myex{\label{ex:enumeration} \actor{Merkel}, \actor{Spahn} und \actor{Gates} are already planning the next attack.}
\myex{\label{ex:illuminaten} This vague ``war on terror'' was apparently used by the \actor{Illuminati} and their \actor{Rothschild puppets} (\actor{Bush} and \actor{Obama}) [...].}

An exception applies to coordination ellipsis constructions, where a shared constituent is omitted from all but the final element of a coordinated list. In such cases, the entire construction is annotated as a single span (\myref{ex:beamte}).

\myex{\label{ex:beamte} \actor{Polizei-, Justiz- und Verwaltungsbeamte}\\\textit{English: \actor{Police}, \actor{judicial}, and \actor{administrative} \actor{officials}}}

\needspace{3\baselineskip}
\paragraph{Adjectival reference.} Attributive adjectives may themselves carry an actor reference if they evoke a known conspiratorial group, independent of the noun they modify. In (\myref{ex:satanic_bible}), \textit{satanic} does not only describe a property of the noun \textit{bible}, but instead invokes the conspiratorial actor group of ``Satanists''. In such cases, only the adjective is annotated as actor, while the noun phrase it modifies is excluded.

\myex{\label{ex:satanic_bible} \actor{satanic} bible}

\subsubsection{Character Level}

\needspace{3\baselineskip}
\paragraph{Sub-word annotation.}\label{rule:sub-word} Spans should usually cover at least one full word. Sub-word annotation is permitted in case of determinative compounds which are typical for German, where the first constituent, i.e. the modifier, carries conspiracy-related reference (\myref{ex:freimaurer}) or where the compound has supposedly been created by a mistake in processing (\myref{ex:dänemark}).

\myex{\label{ex:freimaurer} \actor{Freimaurer}stadt\\\textit{\actor{Freemason} city}}
\myex{\label{ex:dänemark} Im\actor{Denmark}}

In some cases, usually written with a hyphen, the modifier as well as the entire compound relate to a conspiracy theory. In these cases, both the full compound and the modifier constituent are annotated as instances of nested annotation, e.g., \textit{Extinction} as well as \textit{Extinction freak} in (\myref{ex:extinction}) (see also the Nested actors rule).

\myex{\label{ex:extinction} \actor{\actorinner{Extinction}-Irrer}\\\textit{\actor{\actorinner{Extinction} freak}}}

\needspace{3\baselineskip}
\paragraph{Abbreviations.}\label{rule:abbreviation} If the proper name of, e.g., an organization and its abbreviation appear together, we annotate both together (\myref{ex:who_abbreviated}).

\myex{\label{ex:who_abbreviated} \actor{World Health Organization (WHO)}}

\needspace{3\baselineskip}
\paragraph{Punctuation.} A span must neither begin nor end with a space or punctuation mark, unless it is part of an abbreviated word (\myref{ex:punctuation}).

\myex{\label{ex:punctuation} What do you think is the actual intent of the \actor{WHO}?}
\myex{\label{ex:punctuation_abbrev} The \actor{W.H.O.}}

\needspace{3\baselineskip}
\paragraph{Quotation marks.} If a labeled span is surrounded by quotation marks, both quotation marks must be included in the span (\myref{ex:quotation}). If only one quotation mark is adjacent, it must be excluded (\myref{ex:unsere_rasse}).

\myex{\label{ex:quotation} The ``\actor{government}'' is working on [...].}

\needspace{3\baselineskip}
\paragraph{Emojis.} Emojis alone are not annotated.

\needspace{3\baselineskip}
\subsubsection{Special Cases}

\needspace{3\baselineskip}
\paragraph{Pejorative coinages.} Actors are sometimes referred to through pejorative or mocking coinages rather than their proper names. Such codes serve as a heuristic indicator of a malicious actor (\myref{ex:codes_merkel}). In these cases, the entire construction is annotated as a single span, even if the proper name is embedded within a larger word or verbal construction. The made-up verb ``erspahn'' in (\myref{ex:codes_spahn}) sounds like the German verb ``ersparen'', English ``to spare someone''.

\myex{\label{ex:codes_merkel} \actor{Merkill} ($\rightarrow$ Angela Merkel)}
\myex{\label{ex:codes_spahn} Berlin will uns nichts \actor{er`spahn'} ($\rightarrow$ Jens Spahn)\\\textit{Berlin does not want to spare us.}}

\needspace{3\baselineskip}
\paragraph{Ambiguous pronouns.} In German, the words \textit{Sie} and \textit{sie} may refer to a third-person singular or plural (\textit{she, they}; the possible actor or actors) or the formal second-person subject (\textit{you}; the reader). If the referent is ambiguous the pronoun is not included in annotation.

\needspace{3\baselineskip}
\subsubsection{Negative Examples} The following examples illustrate cases where no actor annotation is to be assigned since no conspiracy theory is being propagated.

\vspace{0.4em}
\noindent
\textit{CT is only referenced, i.e. in meta-language or mentioned dismissively:}

\myex{``It is unacceptable to portray a Jewish global conspiracy or Microsoft co-founder Bill Gates as the creator of the coronavirus,'' says Broemme.}
\myex{Half of the posts these days are about politicians who appear to be pedophiles, the Satanic Gotthard Opening, and similar nonsense.}

\textit{Mentioning expressions that frequently occur in conspiracy narratives, but without conspiratorial framing:}

\myex{The World Health Organization (WHO) warns: ``Everyday masks may increase the risk of COVID-19''}
\myex{``Ethnic and religious tensions ... are the most troubling aspect of the reform process in Myanmar,'' Soros said at the Asia Society in New York.}
\myex{YouTube had deleted my latest video about Gates, Rockefeller, Soros, and the upheaval we're currently experiencing.}

\textit{Criticism without conspiracy framing:}

\myex{More than 40,000 Austrians protested yesterday against the current COVID-19 restrictions and the mandatory vaccination announced by the government.}
\myex{Here is a chancellor on the verge of admitting the failure of her insane policies---and the members of the so-called opposition are incapable of capitalizing on this golden opportunity to demand the resignation of Merkel and the federal government.}
\section{Interannotator Agreement}
\label{sec:appendix_iaa}

\begin{table}[h!]
\centering
\small
\begin{tabular}{lr}
\toprule
Metric & Score \\
\midrule
Gamma & 0.75 \\
Soft Gamma & 0.76 \\
Strict F1 & 0.73 \\
Relaxed F1 & 0.79 \\
Overlap F1 & 0.77 \\
\bottomrule
\end{tabular}
\caption{Inter-annotator agreement on 100 independently annotated Telegram posts.}
\label{tab:appendix_iaa}
\end{table}

\section{Actor Extraction}
\label{sec:appendix_actor_extraction}

Table~\ref{tab:model_training} summarizes the evaluated transformer models, the explored hyperparameter search space and the best-performing configuration. Hyperparameters were selected using stratified 5-fold cross-validation with early stopping based on validation strict token F1 (patience = 3 epochs) and a maximum of 20 training epochs.

\begin{table}[h]
\centering
\small
\begin{tabular}{lp{3cm}l}
\toprule
Setting & Search space & Selected \\
\midrule
Models &
\makecell[l]{GBERT-base\\
GBERT-large\\
GELECTRA-base\\
GottBERT-base}
&
GBERT-base \\
Learning rates &
$1,2,3,5 \times 10^{-5}$ & $3\times10^{-5}$ \\
Batch sizes &
4, 8 & 4 \\
Weight decay &
0.01 & 0.01\\
Epochs & max. 20 & 11\\
\bottomrule
\end{tabular}
\caption{Evaluated models and hyperparameter search space.}
\label{tab:model_training}
\end{table}

The final model was evaluated on the held-out test set across five random seeds. The evaluation results are reported in Table~\ref{tab:model_eval}.

\begin{table}[h]
\centering
\small
\begin{tabular}{lrrrr}
\toprule
Metric & Precision & Recall & F1 & F1 Std. \\
\midrule
Token strict   & 0.52 & 0.53 & 0.52 & 0.007 \\
Token relaxed  & 0.65 & 0.66 & 0.65 & 0.006 \\
Token overlap  & 0.71 & 0.58 & 0.64 & 0.006 \\
Word strict    & 0.55 & 0.54 & 0.55 & 0.010 \\
Word relaxed   & 0.66 & 0.65 & 0.65 & 0.008 \\
Word overlap   & 0.69 & 0.57 & 0.63 & 0.007 \\
\bottomrule
\end{tabular}
\caption{Mean evaluation results across five random seeds.}
\label{tab:model_eval}
\end{table}

\section{Schwurbelarchiv}
\label{sec:appendix_schwurbelarchiv}

\begin{figure}[h]
    \centering
    \includegraphics[width=0.99\linewidth]{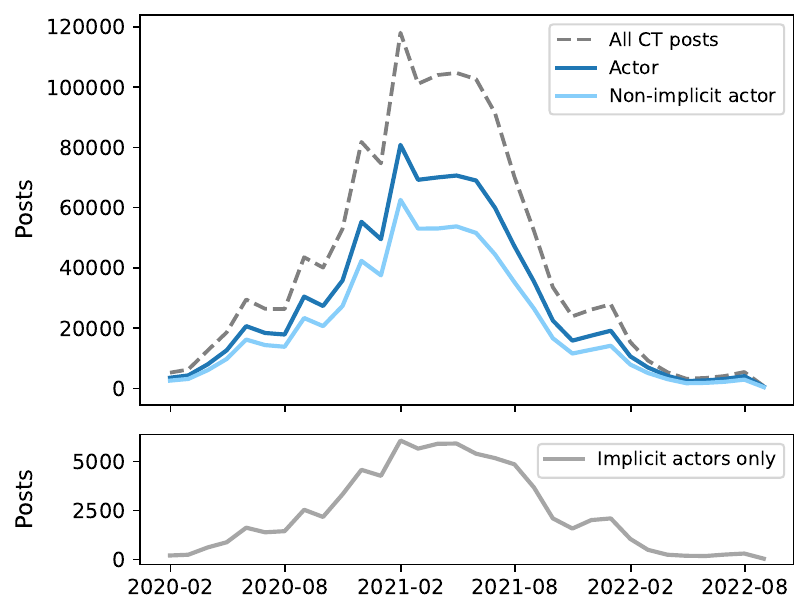}
\caption{Monthly distribution of CT posts and actor annotations in the Schwurbelarchiv corpus from 2020 onwards. The upper panel shows all CT posts, those containing at least one extracted actor, and posts containing at least one annotated non-implicit actor. The lower panel shows CT posts referring exclusively to implicit actors.}   
\label{fig:actor_distribution}
\end{figure}

The Schwurbelarchiv \cite{angermaier_schwurbelarchiv_2026} is a large-scale archive of German-language Telegram channels assembled through snowball sampling starting from known conspiracy-theory channels. The archive comprises approximately 64 million posts between September 2015 and August 2022, 97.6\%  of which were posted from 2020 onward  (see Fig. \ref{fig:actor_distribution}). For our analysis, we use the preprocessed version provided by the authors, which includes document-level conspiracy-theory predictions obtained with the model of \citet{pustet_detection_2024} as well as automatically generated transcriptions of audio and video content. 

To ensure compatibility with our actor extraction model and the  CT document-level classifier, we restrict our analysis to unique non-transcribed posts of at most 2,000 characters that are classified as CT, resulting in $~1.35$ million posts. 
Table~\ref{tab:schwurbelarchiv_filtering} summarizes the filtering steps applied before actor extraction.

\begin{table}[ht!]
\centering
\small
\begin{tabular}{lrr}
\toprule
Filtering step & Posts & Remaining \\
\midrule
Full archive & 63,908,365 & 100.0\% \\
Remove duplicates  & 25,835,928 & 40.4\% \\
Remove AV transcriptions & 25,758,825 & 40.3\% \\
Keep CT posts & 1,397,663 & 2.2\% \\
Limit to $\leq$ 2,000 characters & 1,353,990 & 2.1\% \\
\bottomrule
\end{tabular}
\caption{Filtering pipeline used to obtain the Schwurbelarchiv analysis corpus. }
\label{tab:schwurbelarchiv_filtering}
\end{table}

The resulting corpus contains 1,353,990 conspiracy-theory posts, of which 918,353 (67.8\%) contain at least one extracted actor.
\end{document}